\documentclass[journal]{IEEEtran}

\IEEEoverridecommandlockouts
\usepackage{cite}
\usepackage{pdfrender}
\usepackage{url}

\usepackage{amssymb}

\usepackage{amsmath,amssymb,amsfonts}
\usepackage[noend]{algorithmic}
\usepackage{algorithm}
\usepackage{graphicx}
\usepackage{textcomp}
\usepackage{xcolor}
\usepackage{multirow}
\begin{document}

\title{{\huge CIMNAS: A Joint Framework for Compute-In-Memory-Aware Neural Architecture Search}

\author{Olga Krestinskaya, Mohammed E. Fouda, Ahmed Eltawil, and Khaled N. Salama}
\thanks{Olga Krestinskaya, Ahmed Eltawil and Khaled N. Salama are with King Abdullah University of Science and Technology (KAUST), Thuwal, Saudi Arabia. Emails:\{ok@ieee.org,  ahmed.eltawil@kaust.edu.sa, khaled.salama@kaust.edu.sa\}. Mohammed Fouda is with Compumacy for Artificial Intelligence Solutions, Cairo, Egypt. Email: foudam@uci.edu.}
\thanks{This work was supported by the King Abdullah University of Science and Technology through the Competitive Research Grant program under grant  URF/1/4704-01-01.}
}
\maketitle

\begin{abstract}

To maximize hardware efficiency and performance accuracy in Compute-In-Memory (CIM)-based neural network accelerators for Artificial Intelligence (AI) applications, co-optimizing both software and hardware design parameters is essential.
Manual tuning is impractical due to the vast number of parameters and their complex interdependencies. To effectively automate the design and optimization of CIM-based neural network accelerators, hardware-aware neural architecture search (HW-NAS) techniques can be applied.
This work introduces CIMNAS, a joint model-quantization-hardware optimization framework for CIM architectures. 
CIMNAS simultaneously searches across software parameters, quantization policies, and a broad range of hardware parameters, incorporating device-, circuit-, and architecture-level co-optimizations.
CIMNAS experiments were conducted over a search space of 9.9$\times$10$^{85}$ potential parameter combinations with the MobileNet model as a baseline {\color{black}and RRAM-based CIM architecture}.
Evaluated on the ImageNet dataset, CIMNAS achieved a reduction in energy-delay-area product (EDAP) ranging from $90.1\times$ to $104.5\times$, an improvement in TOPS/W between $4.68\times$ and $4.82\times$, and an enhancement in TOPS/mm$^2$ from $11.3\times$ to $12.78\times$ relative to various baselines, all while maintaining an accuracy of $73.81\%$. {\color{black} The adaptability and robustness of CIMNAS are demonstrated by extending the framework to support the SRAM-based ResNet50 architecture, achieving up to an $819.5\times$ reduction in EDAP.}
Unlike other state-of-the-art methods, CIMNAS achieves EDAP-focused optimization without any accuracy loss, generating diverse software-hardware parameter combinations for high-performance CIM-based neural network designs. The source code of CIMNAS is available at \url{https://github.com/OlgaKrestinskaya/CIMNAS}.

\end{abstract}


\section{Introduction}

The exponential growth of Artificial Intelligence (AI) applications and increasing AI model complexity are raising the energy demands for training and processing AI workloads \cite{mehonic2024roadmap}. This trend has created a demand for more sustainable and energy-efficient hardware solutions for AI applications. 
Compute-In-Memory (CIM) neural network accelerators have emerged as promising architectures for achieving energy-efficient AI processing \cite{sebastian2020memory,krestinskaya2023towards, ielmini2020device, yantir2022hardware, smagulova2023resistive}.
To maximize the hardware efficiency of CIM accelerators and maintain high performance for neural network workloads, it is essential to co-optimize both neural network model parameters and CIM hardware parameters \cite{aguirre2024hardware}. Furthermore, achieving optimal efficiency in CIM hardware requires a holistic approach that considers all levels of hardware design, including device-, circuit-, and architecture-level optimizations \cite{zhang2020neuro}.
With a design space reaching $10^{85}$ possible parameter combinations,
manually optimizing such an extensive design space is infeasible. An effective approach to address this co-optimization challenge is Hardware-Aware Neural Architecture Search (HW-NAS) \cite{krestinskaya2024neural, chitty2022neural,rakka2024review}.

Initially, neural architecture search (NAS) techniques were developed for purely software-based neural network models to automate the search for optimal parameter combinations \cite{ren2021comprehensive}. Later, these techniques incorporated hardware feedback \cite{xu2024nash,chitty2022neural} and were adapted for various hardware types, including CIM architectures \cite{krestinskaya2024neural, guan2022hardware}. Current state-of-the-art HW-NAS frameworks for CIM primarily focus on optimizing neural network models for hardware implementation \cite{benmeziane2023analognas, yuan2021nas4rram, jiang2020device}, performing hardware design space exploration separately \cite{han2024comn, krestinskaya2024neural, yang2021multi}, or jointly optimizing model parameters with a limited set of hardware parameters \cite{negi2022nax, moitra2023xpert}.
However, to achieve both high model accuracy and efficient hardware performance in CIM design, it is essential to co-optimize a broad range of parameters, especially due to their complex interdependencies.


\begin{table*}[t!]
\centering
\caption{State-of-the-art HW-NAS approaches and comparison with CIMNAS.}
\resizebox{2.0\columnwidth}{!}{%
\begin{tabular}{|lcccccccccccc|}
\hline
\multicolumn{1}{|l|}{\multirow{3}{*}{\textbf{Framework}}}                                    & \multicolumn{5}{c|}{\textbf{Optimization (search space)}}                                                                                                                                                                                                           & \multicolumn{1}{c|}{\multirow{3}{*}{\textbf{Algorithm}}} & \multicolumn{1}{c|}{\multirow{3}{*}{\textbf{\begin{tabular}[c]{@{}c@{}}Search \\ space \\ size\end{tabular}}}} & \multicolumn{1}{c|}{\multirow{3}{*}{\textbf{\begin{tabular}[c]{@{}c@{}}SW-HW \\ co-opt.\end{tabular}}}} & \multicolumn{1}{c|}{\multirow{3}{*}{\textbf{Approach}}}                                     & \multicolumn{1}{c|}{\multirow{3}{*}{\textbf{\begin{tabular}[c]{@{}c@{}}Backbone \\ network\end{tabular}}}} & \multicolumn{1}{c|}{\multirow{3}{*}{\textbf{\begin{tabular}[c]{@{}c@{}}$\frac{\text{Baseline EDAP}^{*5}}{\text{Optimized EDAP}}$\end{tabular}}}} 
& 

\multirow{3}{*}{\textbf{\begin{tabular}[c]{@{}c@{}} Accuracy drop \\(EDAP\\- optimized)$^{*6}$ \end{tabular}}}  

\\ \cline{2-6}
\multicolumn{1}{|l|}{}                                                                       & \multicolumn{1}{c|}{\multirow{2}{*}{\textbf{\begin{tabular}[c]{@{}c@{}}SW/\\ Model\end{tabular}}}} & \multicolumn{1}{c|}{\multirow{2}{*}{\textbf{Q$^{*1}$}}} & \multicolumn{3}{c|}{\textbf{Hardware}}                                                                    & \multicolumn{1}{c|}{}                                    & \multicolumn{1}{c|}{}                                                                                          & \multicolumn{1}{c|}{}                                                                                   & \multicolumn{1}{c|}{}                                                                       & \multicolumn{1}{c|}{}                                                                                      & \multicolumn{1}{c|}{}                                                                                                    &                                                                                                             \\ \cline{4-6}
\multicolumn{1}{|l|}{}                                                                       & \multicolumn{1}{c|}{}                                                                              & \multicolumn{1}{c|}{}                              & \multicolumn{1}{c|}{\textbf{D$^{*2}$}} & \multicolumn{1}{c|}{\textbf{C$^{*3}$}} & \multicolumn{1}{c|}{\textbf{A$^{*4}$}} & \multicolumn{1}{c|}{}                                    & \multicolumn{1}{c|}{}                                                                                          & \multicolumn{1}{c|}{}                                                                                   & \multicolumn{1}{c|}{}                                                                       & \multicolumn{1}{c|}{}                                                                                      & \multicolumn{1}{c|}{}                                                                                                    &                                                                                                             \\ \hline
\multicolumn{1}{|l|}{\textbf{AnalogNAS} \cite{benmeziane2023analognas}}                                                     & \multicolumn{1}{c|}{\checkmark}                                                                             & \multicolumn{1}{c|}{-}                             & \multicolumn{1}{c|}{-}            & \multicolumn{1}{c|}{-}            & \multicolumn{1}{c|}{-}            & \multicolumn{1}{c|}{EA}                                  & \multicolumn{1}{c|}{7.3$\times$10$^{10}$}                                                                                   & \multicolumn{1}{c|}{-}                                                                                  & \multicolumn{1}{c|}{HW feedback}                                                            & \multicolumn{1}{c|}{ResNet32}                                                                              & \multicolumn{1}{c|}{-}                                                                                                   & -                                                                                                           \\ \hline
\multicolumn{1}{|l|}{\textbf{\begin{tabular}[c]{@{}l@{}}NAS4RRAM \cite{yuan2021nas4rram}\end{tabular}}}     & \multicolumn{1}{c|}{\checkmark}                                                                             & \multicolumn{1}{c|}{-}                             & \multicolumn{1}{c|}{-}            & \multicolumn{1}{c|}{-}            & \multicolumn{1}{c|}{-}            & \multicolumn{1}{c|}{EA}                                  & \multicolumn{1}{c|}{4.7$\times$10$^{5}$}                                                                                    & \multicolumn{1}{c|}{-}                                                                                  & \multicolumn{1}{c|}{HW feedback}                                                            & \multicolumn{1}{c|}{\begin{tabular}[c]{@{}c@{}}ResNet20,\\ ResNet32\end{tabular}}                           & \multicolumn{1}{c|}{-}                                                                                                   & -                                                                                                           \\ \hline
\multicolumn{1}{|l|}{\textbf{Flash} \cite{li2021flash}}                                                         & \multicolumn{1}{c|}{\checkmark}                                                                             & \multicolumn{1}{c|}{-}                             & \multicolumn{1}{c|}{-}            & \multicolumn{1}{c|}{-}            & \multicolumn{1}{c|}{-}            & \multicolumn{1}{c|}{SHGO}                                & \multicolumn{1}{c|}{6.4$\times$10$^{10}$}                                                                                   & \multicolumn{1}{c|}{-}                                                                                  & \multicolumn{1}{c|}{HW feedback}                                                           & \multicolumn{1}{c|}{DenseNets}                                                                             & \multicolumn{1}{c|}{-}                                                                                                   & -                                                                                                           \\ \hline

\multicolumn{1}{|l|}{\textbf{NACIM} \cite{jiang2020device}}                                                         & \multicolumn{1}{c|}{\checkmark}                                                                             & \multicolumn{1}{c|}{\checkmark}                             & \multicolumn{1}{c|}{-}            & \multicolumn{1}{c|}{-}            & \multicolumn{1}{c|}{-}            & \multicolumn{1}{c|}{RL}                                  & \multicolumn{1}{c|}{2.6$\times$10$^{25}$}                                                                                   & \multicolumn{1}{c|}{-}                                                                                  & \multicolumn{1}{c|}{HW feedback}                                                            & \multicolumn{1}{c|}{VGG11}                                                                                 & \multicolumn{1}{c|}{3.9}                                                                                                 & $\downarrow$ 11.0\%                                                                                                     \\ \hline

\multicolumn{1}{|l|}{\textbf{CoMN}$^{*7}$ \cite{han2024comn}}                                                          & \multicolumn{1}{c|}{-}                                                                             & \multicolumn{1}{c|}{-}                             & \multicolumn{1}{c|}{\checkmark}            & \multicolumn{1}{c|}{\checkmark}            & \multicolumn{1}{c|}{\checkmark}            & \multicolumn{1}{c|}{BO}                                  & \multicolumn{1}{c|}{2.2$\times$10$^4$}                                                                                  & \multicolumn{1}{c|}{-}                                                                                  & \multicolumn{1}{c|}{\begin{tabular}[c]{@{}c@{}}HW feedback, \\ DSE, two-stage\end{tabular}} & \multicolumn{1}{c|}{CNNs$^{*8}$}                                                                               & \multicolumn{1}{c|}{-}                                                                                                   & -                                                                                                           \\ \hline
\multicolumn{1}{|l|}{\textbf{\begin{tabular}[c]{@{}l@{}}Joint HWC$^{*9}$ \end{tabular}} \cite{krestinskaya2024towards}}  & \multicolumn{1}{c|}{-}                                                                             & \multicolumn{1}{c|}{-}                             & \multicolumn{1}{c|}{\checkmark}            & \multicolumn{1}{c|}{\checkmark}            & \multicolumn{1}{c|}{\checkmark}            & \multicolumn{1}{c|}{EA}                                  & \multicolumn{1}{c|}{1.9$\times$10$^{7}$}                                                                                    & \multicolumn{1}{c|}{-}                                                                                  & \multicolumn{1}{c|}{DSE}                                                                    & \multicolumn{1}{c|}{CNNs$^{*8}$}                                                                               & \multicolumn{1}{c|}{-}                                                                                                   & -                                                                                                           \\ \hline
\multicolumn{1}{|l|}{\textbf{NAX} \cite{negi2022nax}}                                                           & \multicolumn{1}{c|}{\checkmark}                                                                             & \multicolumn{1}{c|}{-}                             & \multicolumn{1}{c|}{-}            & \multicolumn{1}{c|}{\checkmark}            & \multicolumn{1}{c|}{-}            & \multicolumn{1}{c|}{DS}                                  & \multicolumn{1}{c|}{2.4$\times$10$^{11}$}                                                                                   & \multicolumn{1}{c|}{\checkmark}                                                                                  & \multicolumn{1}{c|}{Joint}                                                                  & \multicolumn{1}{c|}{ResNet20}                                                                              & \multicolumn{1}{c|}{1.1-5.9$^{*10}$}                                                                                          & $\downarrow$ 1.1-11.1\%$^{*10}$                                                                                                 \\ \hline

\multicolumn{1}{|l|}{{\color{black}\textbf{CIMNet} \cite{chen2024cimnet}}}                                                        & \multicolumn{1}{c|}{\color{black}{\checkmark}}                                                                             & \multicolumn{1}{c|}{{\color{black}\checkmark}}                             & \multicolumn{1}{c|}{{\color{black}\checkmark}}            & \multicolumn{1}{c|}{{\color{black}\checkmark}}            & \multicolumn{1}{c|}{{\color{black}-}}            & \multicolumn{1}{c|}{{\color{black}EA}}                                  & \multicolumn{1}{c|}{{\color{black}1.0$\times$10$^{12}$} }                                                                                  & \multicolumn{1}{c|}{{\color{black}\checkmark}}                                                                                  & \multicolumn{1}{c|}{{\color{black}Joint}}                                                                  & \multicolumn{1}{c|}{{\color{black}EfficientNet}}                                                                              & \multicolumn{1}{c|}{{\color{black}-}}                                                                                      & {\color{black}$\uparrow$ 0.1-0.2\%$^{*12}$}                                                                                              \\ \hline

\multicolumn{1}{|l|}{\textbf{Gibbon} \cite{sun2023gibbon}}                                                        & \multicolumn{1}{c|}{\checkmark}                                                                             & \multicolumn{1}{c|}{\checkmark}                             & \multicolumn{1}{c|}{\checkmark}            & \multicolumn{1}{c|}{\checkmark}            & \multicolumn{1}{c|}{-}            & \multicolumn{1}{c|}{EA}                                  & \multicolumn{1}{c|}{4.3$\times$10$^{84}$}                                                                                   & \multicolumn{1}{c|}{\checkmark}                                                                                  & \multicolumn{1}{c|}{Joint}                                                                  & \multicolumn{1}{c|}{ResNet18}                                                                              & \multicolumn{1}{c|}{51.1-59.1$^{*10}$}                                                                                      & $\downarrow$ 6.3-6.7\%$^{*10}$                                                                                              \\ \hline
\multicolumn{1}{|l|}{\textbf{XPert} \cite{moitra2023xpert}}                                                         & \multicolumn{1}{c|}{\checkmark}                                                                             & \multicolumn{1}{c|}{\checkmark$^{*11}$}                          & \multicolumn{1}{c|}{-}            & \multicolumn{1}{c|}{\checkmark}            & \multicolumn{1}{c|}{-}            & \multicolumn{1}{c|}{DS}                                  & \multicolumn{1}{c|}{7.1$\times$10$^{34}$}                                                                                   & \multicolumn{1}{c|}{\checkmark}                                                                                  & \multicolumn{1}{c|}{Two-stage}                                                              & \multicolumn{1}{c|}{VGG16}                                                                                 & \multicolumn{1}{c|}{10.4}                                                                                                & $\downarrow$ 1.2-1.3\%                                                                                                       \\ \hline

\multicolumn{1}{|l|}{\multirow{2}{*}{\textbf{\begin{tabular}[c]{@{}l@{}}CIMNAS \\ (this work)\end{tabular}}}}

& \multicolumn{1}{c|}{\multirow{2}{*}{\checkmark}} 
& \multicolumn{1}{c|}{\multirow{2}{*}{\checkmark}}                             & \multicolumn{1}{c|}{\multirow{2}{*}{\checkmark}}            & \multicolumn{1}{c|}{\multirow{2}{*}{\checkmark}}            & \multicolumn{1}{c|}{\multirow{2}{*}{\checkmark}}            & \multicolumn{1}{c|}{\multirow{2}{*}{EA}}                                  & \multicolumn{1}{c|}{\textbf{9.9}$\times$\textbf{10}$^{\textbf{85}}$}                                                                                   & \multicolumn{1}{c|}{\textbf{\multirow{2}{*}{\checkmark}}}                                                                                  & \multicolumn{1}{c|}{\multirow{2}{*}{Joint} }                                                                 & \multicolumn{1}{c|}{MobileNet}                                                                             & \multicolumn{1}{c|}{\begin{tabular}[c]{@{}c@{}} \textbf{73.0-92.7}$^{*13}$ \\ \textbf{82.4-107.5}$^{*14}$ \end{tabular}}

& $\uparrow$ \textbf{0.6\%}$^{*15}$- \textbf{0.8\%}$^{*16}$                                                                       
\\ \cline{8-8} \cline{11-13}
\multicolumn{1}{|l|}{} & \multicolumn{1}{c|}{}                                                                             & \multicolumn{1}{c|}{}                             & \multicolumn{1}{c|}{}            & \multicolumn{1}{c|}{}            & \multicolumn{1}{c|}{}            & \multicolumn{1}{c|}{}                                  & \multicolumn{1}{c|}{{\color{black}{1.02}$\times${10}$^{{16}}$} }                                                                                  & \multicolumn{1}{c|}{}                                                                                  & \multicolumn{1}{c|}{}                                                                  & \multicolumn{1}{c|}{{\color{black}ResNet50}}                                                                             & 
\multicolumn{1}{c|}{{\color{black}\begin{tabular}[c]{@{}c@{}} \textbf{191.8-251.1}$^{*13}$ \\ \textbf{622.9-819.5}$^{*14}$ \end{tabular}}}

& {\color{black}$\downarrow$ \textbf{0.6\%}$^{*15}$- \textbf{0.7\%}$^{*16}$} 

\\ \hline
\multicolumn{13}{|l|}{\begin{tabular}[c]{@{}l@{}}SW - software, HW - hardware, DSE - design space exploration, co-opt. - co-optimization. $^{*1}$: Quantization optimization: optimum weights and input precision search.\\ $^{*2}$: Device optimization - including device parameters search, e.g. bits per cell. $^{*3}$: Circuit optimization - including circuit-level parameters optimization of crossbar macros, e.g. \\ array  size or ADC precision.$^{*4}$: Architecture optimization - including higher-level architecture parameters optimization, e.g. tiles, global buffer, etc. $^{*5}$: higher = better, \\$^{*6}$: $\downarrow$ - accuracy decreased, $\uparrow$ - accuracy increased,
$^{*7}$:
Search space size is based on demonstrated HW search space,  $^{*8}$: CNNs: ResNet, VGG,  AlexNet,  MobileNet, \\  $^{*9}$ Hardware-workload co-exploration (HWC), $^{*10}$: varies depending on the dataset, $^{*11}$: input precision only,  
{\color{black}$^{*12}$: energy-latency optimized (area not considered),}  \\{\color{black} $^{*13}$/$^{*14}$: comparing to Baseline 1 / Baseline 2 for 5 best EDAP-optimized designs (Section \ref{results}),
$^{*15}$/$^{*16}$: top1/top-5  performance   accuracies of EDAP-optimized designs versus}  \\ {\color{black}  accuracies of baseline architectures (Section \ref{results}) }\\ \end{tabular}}                                                                                                                                                                                                                                               \\ \hline
\end{tabular}
}
\label{t1}

\end{table*}

In this work, we introduce CIMNAS, a Compute-In-Memory-Aware Neural Architecture Search framework that performs joint co-optimization of model parameters, quantization policies, and CIM hardware parameters, with a focus on optimizing a wide range of hardware parameters to enhance efficiency without sacrificing performance accuracy. 
Unlike traditional approaches, CIMNAS targets a comprehensive range of hardware-level parameters, including device, circuit, and architectural levels, to maximize hardware efficiency without compromising model accuracy.
CIMNAS explores an extensive search space of size 9.9$\times$10$^{85}$ of possible combinations of parameters, jointly optimizing neural network model parameters (e.g. number of layers, kernel sizes, and per-layer expansion factors), quantization policies (e.g. layer-wise precision for weights and activations for depth-wise and point-wise convolutions), and CIM hardware configurations (e.g. bits per cell in CIM device, operation voltage, cycle time, CIM crossbar sizes, number of CIM macros per tile, number of tiles sharing a single router, number of tile groups per chip and global buffer size).
By integrating all these dimensions into a joint search process, CIMNAS avoids suboptimal solutions caused by independent or sequential tuning and mitigates the risk of local minima.
In contrast to state-of-the-art frameworks, CIMNAS is uniquely tailored for the full CIM design hierarchy, enabling the discovery of EDAP-optimized (Energy–Delay–Area–Product) hardware configurations while maintaining competitive accuracy. 
Moreover, CIMNAS is robust, adaptable, and offers a high diversity among top-optimized designs, acknowledging the potential for multiple optimal parameter configurations within such a vast search space.

This paper is structured as follows. Section~\ref{s2} provides background on Compute-In-Memory (CIM) architectures and reviews related work in neural architecture search and hardware-aware optimization. Section~\ref{s3} presents the proposed CIMNAS framework, describing its joint search space formulation, optimization methodology, hardware configuration, neural network model used in the simulations, and the evaluation metrics for both software and hardware. Section~\ref{results} details the experimental setup, including search configurations, baseline models, and comparison methods, and provides an in-depth analysis of the results, highlighting CIMNAS’s performance in terms of accuracy, efficiency, and design diversity. Section~\ref{discussion} emphasizes the importance of jointly optimizing model architecture, quantization policies, and hardware parameters, and outlines how CIMNAS can be adapted for future co-design applications and emerging CIM technologies. Finally, Section~\ref{conclusion} concludes the paper and outlines potential directions for future research.

\section{Related Work}
\label{s2}

State-of-the-art HW-NAS methods for CIM-based architectures can be categorized into three main approaches: HW-NAS for fixed CIM architectures, CIM-based design space exploration for a fixed optimized neural network model, and HW-NAS for co-optimization of software-hardware parameters in CIM-based neural network designs \cite{krestinskaya2024neural}. In HW-NAS for fixed CIM architectures, neural network model parameters are optimized with consideration for CIM hardware (HW) feedback \cite{benmeziane2023analognas, krestinskaya2020automating, jiang2020device, guan2022hardware, yan2021uncertainty,  yuan2021nas4rram, li2021flash}. 
This approach adapts the neural network model to fit the fixed hardware, helping to mitigate the effects of hardware non-idealities in CIM devices \cite{krestinskaya2020towards}
Similarly, several frameworks optimize quantization policies to enhance the performance of CIM-based architectures \cite{huang2021mixed, kang2021genetic, peng2022cmq}. 
In contrast, CIM-based design space exploration focuses on identifying the optimal CIM hardware parameters for deploying a fixed neural network model \cite{yang2021multi, krestinskaya2024towards, han2024comn}.
Finally, HW-NAS methods that co-optimize software and hardware parameters focus on jointly optimizing both the neural network model and CIM hardware parameters \cite{sun2023gibbon, negi2022nax, moitra2023xpert, chen2024cimnet}. This approach is particularly valuable in the initial stages of IMC-based AI accelerator design, where an optimized model and hardware setup are required for specific applications.

Table \ref{t1} presents a comparison between the most relevant state-of-the-art CIM-based HW-NAS approaches and the proposed CIMNAS framework, evaluating them in terms of optimized parameters, algorithms, search space size, optimization approach, and achieved optimization results. The energy-delay-area product (EDAP) metric is used to compare co-optimization frameworks. Due to variations in datasets and baseline CIM hardware across different frameworks, a direct quantitative comparison with the proposed approach is not feasible. To address this, we use two comparison metrics: (1) the ratio of the EDAP of the baseline design to that of the optimized design, $\frac{Baseline \, EDAP}{Optimized \, EDAP}$ \cite{moitra2023xpert}, and (2) the accuracy drop observed in the EDAP-optimized design. {\color{black} Baseline EDAP refers to the EDAP of the reference design used as the starting point in the framework. Optimized EDAP denotes the EDAP of the best-performing architecture obtained after optimizing hardware parameters. Accuracy drop indicates the reduction in performance accuracy of the optimized design relative to the baseline, when optimized to reduce EDAP.} These metrics are presented in Table \ref{t1} to provide a fair and consistent evaluation.  Accuracy drop is commonly observed in software-hardware co-optimization, where accuracy is often traded off to achieve greater hardware efficiency.

\begin{figure*}[t!]
    \includegraphics[width=\textwidth]{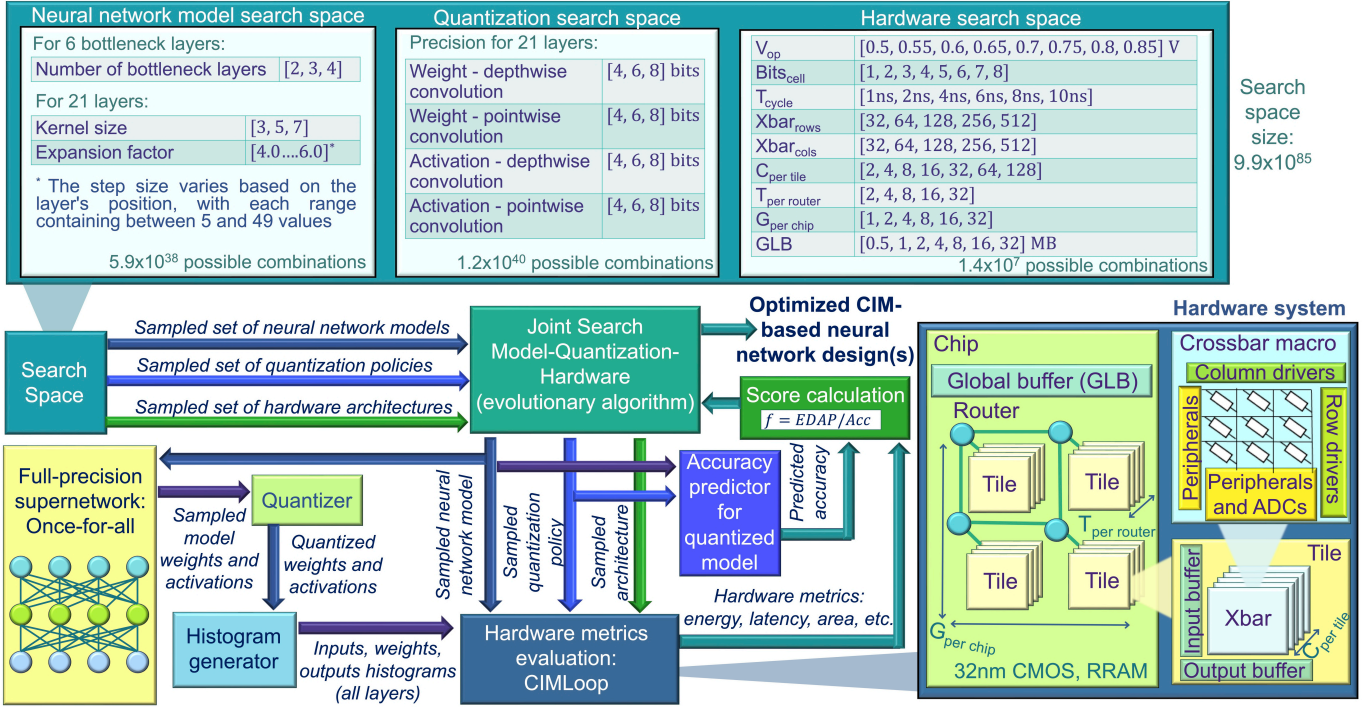}
    \caption{CIMNAS: Joint co-optimization of software, precision, and hardware parameters for CIM-based neural network design.}
    \label{f2}
\end{figure*}

AnalogNAS \cite{benmeziane2023analognas}, NAS4RRAM \cite{yuan2021nas4rram}, Flash \cite{li2021flash}, and NACIM \cite{jiang2020device} frameworks focus on HW-NAS for optimizing neural network models with hardware feedback. Both AnalogNAS and NAS4RRAM optimize ResNet-like models using evolutionary algorithms (EA). Flash employs a coarse- and fine-grained search via a simplicial homology global optimization (SHGO)-based algorithm, utilizing an accuracy predictor and neural network degree metrics \cite{li2021flash}. NACIM also optimizes model precision alongside parameters using reinforcement learning (RL) \cite{jiang2020device}. These approaches optimize neural network models for deployment on existing CIM-based hardware.
In contrast, CoMN \cite{han2024comn} and the Joint hardware-workload co-optimization (HWC) framework \cite{krestinskaya2024towards} focus on hardware design space exploration (DSE) for fixed neural network models. CoMN employs Bayesian optimization (BO) to search a large hardware space, optimizing mapping and architecture parameters for system performance. The HWC framework performs joint optimization across hardware workloads to create a generalized hardware solution optimized for diverse tasks. However, separately optimizing hardware parameters for high-accuracy models may result in suboptimal designs, as software-optimized models often lead to underutilized CIM-based hardware in deployment \cite{sun2023gibbon}.

Achieving truly optimal CIM chip design for neural network applications, particularly for low-power edge devices, requires co-optimizing software and hardware parameters. Frameworks like NAX \cite{negi2022nax}, {\color{black} CIMNet \cite{chen2024cimnet},} Gibbon \cite{sun2023gibbon}, and XPert \cite{moitra2023xpert}. illustrate such co-optimization approaches. NAX optimizes kernel and crossbar sizes for energy efficiency using differential search (DS), but focuses only on limited crossbar array trade-offs, which is insufficient for full CIM architecture optimization.
{\color{black} CIMNet jointly optimizes model architecture, quantization levels, and hardware parameters by leveraging an accuracy predictor and a layer-wise look-up table (LUT)-based hardware metrics estimator, assuming a constrained hardware search space limited to device precision and crossbar size. While this approach significantly reduces search time, the reliance on layer-wise LUT estimators becomes impractical as the hardware search space expands, particularly when architectural and system-level parameters are included. These parameters often depend on inter-layer data transmission and mapping strategies, making LUT-based estimation inefficient and less scalable. }
Gibbon co-optimizes a large search space including software precision and CIM hardware parameters, yet restricts exploration to crossbar macro settings, such as crossbar and converter precision, leading to a notable accuracy drop due to modifications in ResNet for better CIM compatibility. To reduce the search space from 4.3$\times$10$^{84}$ to 1.3$\times$10$^{22}$, Gibbon’s algorithm prunes lower-priority parameters. Applying a similar strategy to the search space in this work could result in missed optimal configurations due to strong interdependencies among parameters such as circuit settings, layer count, and precision.
XPert employs a two-stage optimization, first optimizing channel depth, ADC type, and column sharing to reduce latency and area, then adjusting input and ADC precision for energy and accuracy. Although effective for initial parameters, this approach may get trapped in local minima, limiting the diversity of designs. Both Gibbon and XPert vary hardware parameters across network layers, which, while enhancing efficiency, complicates fabrication and limits the chip’s reusability for other tasks due to inconsistencies like differing crossbar and ADC designs.

In this work, we address these issues by jointly co-optimizing software model parameters (e.g. number of layers, kernel sizes, expansion factor for each layer), quantization settings (e.g. precisions for weights and activations for depth-wise and point-wise convolution in each layer), and hardware parameters, including device-, circuit-, and architecture-level configurations, within a vast search space of 9.9$\times$10$^{85}$ possible combinations of parameters. We explore diverse hardware parameters while maintaining consistency across the architecture to facilitate easier transistor-level design, layout, and CIM chip fabrication based on these optimized configurations.
We jointly co-optimize these parameters within a single search to avoid local minima, fully explore the search space, and ensure diversity among the optimized designs generated by the framework. CIMNAS preserves the high performance accuracy of CIM-based neural networks while achieving optimized EDAP, avoiding a drop in accuracy relative to the baseline design.

\section{CIMNAS}
\label{s3}

The proposed CIMNAS framework for joint software-quantization-hardware co-exploration in CIM-based neural network optimization is illustrated in Fig. \ref{f2}. The CIMNAS search space is built upon the MobileNetV2 baseline architecture \cite{sandler2018mobilenetv2} for an ImageNet-based \cite{deng2009imagenet} image classification task. The MobileNet backbone was used as it is often overlooked in in-memory computing research and poses challenges due to its depthwise separable convolutions and compact design \cite{zhou2021analognets}. Despite these complexities, our results demonstrate that our algorithm performs effectively, highlighting its robustness and adaptability.
To avoid local minima and ensure comprehensive search space coverage, CIMNAS jointly optimizes all parameters in the search space. The framework samples sets of neural network parameters, quantization policies, and hardware architectures, feeding them into an evolutionary joint search algorithm. In each iteration, the algorithm evaluates each neural network implementation based on performance accuracy and hardware metrics. An accuracy predictor, trained on quantized neural network models, is used to quickly estimate performance accuracy, as quantizing and retraining for each combination is impractical. 
{\color{black} Hardware metrics are evaluated using sampled hardware parameters and layer-specific histograms of quantized inputs, weights, and outputs. These histograms are generated by sampling a candidate neural network from a pre-trained full-precision supernetwork and applying the corresponding sampled quantization policy (Section \ref{sec:C}).} CIMNAS outputs a set of optimized CIM-based mixed-precision neural network models and hardware parameters. The algorithm and overall workflow of CIMNAS are presented in Algorithm \ref{alg}.

\begin{algorithm}[ht]
    \caption{CIMNAS algorithm (in this work $P=150$, $G=70-100$).}\label{alg}
    \begin{algorithmic}
        \STATE  $\mathit{Initial \, population \, sampling:}$
        \WHILE{$(\mathrm{population} \, p < P)$}
            \STATE $\mathrm{Randomly \,\mathbf{sample} \,neural \, network\,model \,}  \mathrm{\,from\,} S_M$
            \STATE $\mathrm{Randomly \,\mathbf{sample} \,quantization \,policy \,} \mathrm{\,from\,} S_Q$
            \STATE $\mathrm{Randomly \,\mathbf{sample} \,candidate \,hardware \, }  c_H \, \mathrm{\,from\,} S_H$
            \STATE  $\mathbf{Find} \, M \, \mathrm{memory \, \,elements \,required \, for\, the\, sampled} $
            \STATE  $ \qquad \mathrm{\,quantized \,model} $
            \IF{$\mathrm{Number \, of  \,memory \, elements \, of \, } c_h \, \geq M$}
                \STATE $\mathrm{\mathbf{Keep} \, the \, sample \, in\, the\, initial \,population}$
                \STATE $p=p+1$
            \ENDIF
       \ENDWHILE
        \FOR{$g \, \mathbf{in} \, G \,\mathrm{generations}$}
            \STATE  $\mathit{Evaluation \, phase:}$
            \IF{$g>1$}
                \STATE  $\mathbf{Exclude} \, \mathrm{samples, \, where \,hardware \, design \,does \, not }$
                \STATE  $\quad  \mathrm{ fit\, corresponding \, sampled \, model \, and  \, quantization } $
            \ENDIF
            \FOR{$\mathrm{Each \,sample} \, \alpha \, \mathrm{in \, a \, population} $}
                \STATE $\mathrm{\mathbf{Obtain} \,full\, precision\,model \, weights \, and \, activations}$
                \STATE $\qquad \mathrm{ from\, the\, supernetwork}$
                \STATE $\mathrm{\mathbf{Quantize} \, weights \, and \, activation \, values}$
                \STATE $\mathrm{\mathbf{Generate} \, corresponding \, histograms}$
                \STATE $\mathrm{\mathbf{Evaluate} \, hardware \, metrics}$
                \STATE $\mathrm{\mathbf{Obtain} \, accuracy  \, for \, quantized \, model \, from }$
                \STATE $\qquad \mathrm{accuracy  \, predictor}$
                \STATE $\mathrm{\mathbf{Calculate} \, score\,} f_{\alpha}$
            \ENDFOR
            \STATE  $\mathit{Selection, \, crossover, \, and\, mutation:}$
            \STATE $\mathrm{\mathbf{Sort}\, the  \,designs }$
            \STATE $\mathrm{\mathbf{Select} \, designs \, to \, participate\,in \, crossover \, (crossover\,}$
            \STATE $ \qquad \mathrm{ probability\,} \mathbb{P}_c \mathrm{)}$
            \STATE $\mathrm{\mathbf{Perform} \,crossover \,with  \,distribution \, \eta_c, \, constructing}$
            \STATE $\qquad \mathrm{new \, ''offsprings''}$
            \STATE $\mathrm{\mathbf{Execute} \,mutation \,with  \,probability} \, \mathbb{P}_m \mathrm{\, and}$
            \STATE $ \qquad \mathrm{distribition \, \eta_m, \, constructing \, new \, population} $
            \STATE $\qquad \mathrm{ with \,} P \mathrm{\, samples}$
        \ENDFOR
    \end{algorithmic}
\end{algorithm}

\subsection{Combined model-quantization-hardware search space}

The complete search space is illustrated in Fig.~\ref{f2}.
It consists of the neural network model ($S_M$), quantization ($S_Q$), and hardware parameters $S_H$ search spaces. The neural network model search space includes 5.9$\times$10$^{38}$ possible configurations, adjusting parameters such as the number of bottleneck layer blocks (depth/repetition) in the MobileNetV2 architecture, kernel sizes for depthwise convolutions, and expansion factors across six bottleneck layers, while keeping the first convolution, first bottleneck, last convolution, and final linear classification layers fixed.
The quantization search space comprises 1.2$\times$10$^{40}$ possible combinations, covering weight and input precision for depthwise and pointwise convolutions across all bottleneck layers.

The CIM hardware search space is based on a hierarchical resistive random access memory (RRAM)-based CIM architecture, as shown in Fig. \ref{f2}. This hierarchy includes crossbar macros, CIM tiles, tile groups, and the interconnections between them. The CIM chip is composed of crossbar macros with peripheral circuits, row/column drivers, and converters. Each tile consists of multiple crossbar macros ($\mathrm{C_{per \, tile}}$) with $\mathrm{Xbar_{rows}}$ rows and $\mathrm{Xbar_{cols}}$ columns, along with input/output buffers. The chip contains a global SRAM-based buffer ($\mathrm{GLB}$) for storing input and output data, with groups of tiles connected by routers. The CIM chip architecture includes $\mathrm{G_{per \, chip}}$ tile groups, each with $\mathrm{T_{per \, router}}$ tiles and $\mathrm{G_{per \, chip}}$ routers.
All of the parameters mentioned above are incorporated into the search space.
Additional parameters like device precision ($\mathrm{Bits_{cell}}$), operating voltage ($\mathrm{V_{op}}$), and cycle time ($\mathrm{T_{cycle}}$) are also considered, resulting in a hardware search space with 1.4$\times$10$^{7}$ possible parameter combinations.

\begin{figure}[t!]
    \centering
    \includegraphics[width=\columnwidth]{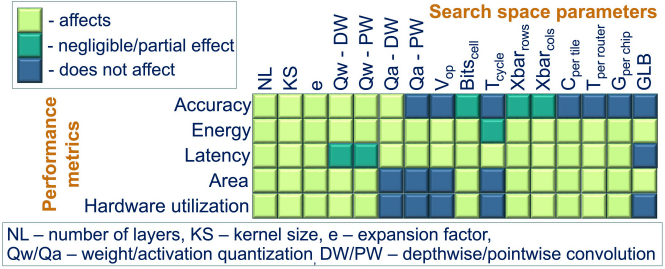}
    \caption{Effect of search space parameters on hardware performance and accuracy and correlation between them.}
    \label{f3}
\end{figure}

In total, the jointly optimized search space includes 9.9$\times$10$^{85}$ possible configurations. 
The relationships between search space parameters and their impact on performance metrics are illustrated in Fig. \ref{f3}, showing strong interdependencies among software, quantization, and hardware parameters. For instance, the sizes of crossbar macros, the number of macros, and the number of tiles are interrelated, allowing for multiple optimal design configurations; reducing one parameter may require increasing another to ensure compatibility between the model and hardware.
Additionally, crossbar size and the number of bits per cell can impact performance accuracy in the presence of device non-idealities and noise variations; however, this topic is beyond the scope of this work.
Overall, most parameters influence all performance metrics, making joint optimization essential to avoid local minima. Furthermore, in Section \ref{results}, we show that two-stage approaches tend to fall into local minima within a given search space, resulting in designs that prioritize either high accuracy or low EDAP, but not both.

\begin{table*}[t!]
\label{ta_algorithms}
\centering
{\color{black}{ 
\caption{Comparison of the state-of-the-art NAS algorithms and advantages of EA.}
\resizebox{2.0\columnwidth}{!}{%
\begin{tabular}{|lcccc|}
\hline
\multicolumn{1}{|l|}{Algorithm}                                                                                               & \multicolumn{1}{c|}{\textbf{RL}}                                                                               & \multicolumn{1}{c|}{\textbf{BO}}                                                                                                      & \multicolumn{1}{c|}{\textbf{DS}}                                                                                                         & \textbf{EA}                                                                                                      \\ \hline
\multicolumn{1}{|l|}{\begin{tabular}[c]{@{}l@{}}Computational \\ complexity\end{tabular}}                                     & \multicolumn{1}{c|}{$O(a^n \cdot t \cdot C)$}                                                                                 & \multicolumn{1}{c|}{$O(k\cdot C+M(k,n))$}                                                                                                    & \multicolumn{1}{c|}{$O(D^{n_d+1}\cdot C$)}                                                                                        & $O(g \cdot p \cdot C)$                                                                                                         \\ \hline
\multicolumn{1}{|l|}{\begin{tabular}[c]{@{}l@{}}Scaling with \\ hyperparameters\end{tabular}}                                 & \multicolumn{1}{c|}{Exponential}                                                                               & \multicolumn{1}{c|}{\begin{tabular}[c]{@{}c@{}}Exponential\\ (surrogate degrades with high $n$)\end{tabular}}                 & \multicolumn{1}{c|}{\begin{tabular}[c]{@{}c@{}}Linear-moderate\\ (memory overhead \\ increases with $n$)\end{tabular}}                 & 
\multicolumn{1}{c|}{\begin{tabular}[c]{@{}c@{}}\textbf{Linear} \textbf{per generation} \\ (robust to large $n$) \end{tabular}} \\ \hline

\multicolumn{1}{|l|}{Parallelization}                                                                                       & \multicolumn{1}{c|}{\begin{tabular}[c]{@{}c@{}}Poor \\ (sequential episodes)\end{tabular}}                     & \multicolumn{1}{c|}{\begin{tabular}[c]{@{}c@{}}Moderate \\ (if batch BO used)\end{tabular}}                                           & \multicolumn{1}{c|}{\begin{tabular}[c]{@{}c@{}}Poor \\ (single graph)\end{tabular}}                                                      & \begin{tabular}[c]{@{}c@{}}\textbf{Good} \\ (population-level parallelism)\end{tabular}                  \\ \hline
\multicolumn{1}{|l|}{\begin{tabular}[c]{@{}l@{}}GPU memory \\ requirement\end{tabular}}                                       & \multicolumn{1}{c|}{\begin{tabular}[c]{@{}c@{}}High \\ (train agents and models)\end{tabular}}                 & \multicolumn{1}{c|}{\begin{tabular}[c]{@{}c@{}}\textbf{Moderate} \\ (depends on M(k,n) and batches)\end{tabular}}                     & \multicolumn{1}{c|}{\begin{tabular}[c]{@{}c@{}}Very high \\ (to store and train supernet)\end{tabular}}                                  & \textbf{Moderate}                                                                                                \\ \hline
\multicolumn{1}{|l|}{\begin{tabular}[c]{@{}l@{}}Applicability to large \\ discrete search space\end{tabular}}                 & \multicolumn{1}{c|}{Hard to scale}                                                                             & \multicolumn{1}{c|}{\begin{tabular}[c]{@{}c@{}}Not applicable \\ (surrogate models breaks down)\end{tabular}}                         & \multicolumn{1}{c|}{\begin{tabular}[c]{@{}c@{}}Not effective \\ for discrete search space\end{tabular}}                                         & \textbf{Applicable}                                                                                              \\ \hline
\multicolumn{5}{|l|}{\begin{tabular}[c]{@{}l@{}}$n$ – number of hyperparameters; $a$ – number of actions per reinforcement learning (RL) step; $t$ – trajectory length (i.e., the number of decisions \\  made   per architecture, approximately equal to $n$); $C$ – cost of evaluating one candidate architecture; $k$ – number of candidates to evaluate;  \\ $M(k,n)$ – complexity  of the surrogate model as a function of $k$ and $n$; $g$ – number of generations in evolutionary algorithms (EA);\\ $p$ – population size in EA; $n_d$ – number of discrete hardware parameters (e.g., hardware-specific parameters that require architecture duplication).\end{tabular}} \\ \hline
\end{tabular}
}}}
\end{table*}

\subsection{Search algorithm and optimization function}

\subsubsection{{\color{black} Search algorithm selection}}

CIMNAS is based on a genetic algorithm, a specific type of evolutionary algorithm (EA). Common {\color{black} NAS and} HW-NAS algorithms include EA, differential search (DS), reinforcement learning (RL), and Bayesian optimization (BO). {\color{black} Table II  compares EA with RL, BO, and DS, justifying the choice of EA for the large and complex CIMNAS search space.

RL is computationally expensive and scales poorly with hyperparameters due to its sequential controller updates. RL-based NAS frameworks NASNet and MnasNet require 22,400 GPU-hours and 288 TPU days, respectively \cite{zoph2018learning, tan2019mnasnet}. RL typically involves 4,000–20,000 episodes per architecture and struggles with sparse rewards in large spaces, leading to suboptimal convergence.
BO is also sequential, where each iteration involves architecture generation, evaluation, and surrogate model updates. It often requires hundreds of surrogate samples (e.g., 283.2 GPU-hours in the BANANAS framework \cite{white2021bananas}), but performance degrades in high-dimensional search spaces due to the breakdown of surrogate modeling. Compared to RL- and BO-based methods, EA is faster and more practical for large search spaces \cite{krestinskaya2024neural, krestinskaya2024towards}.

DS, as in the DARTS framework, trains a supernetwork over a few days \cite{liu2018darts}, but relies on differentiable parameters. When extended to discrete hardware parameters, model duplication leads to exponential memory demands, and relaxation techniques fail to scale.
DS is unsuitable for joint software-hardware searches, as the supernetwork of software models alone requires at least 48GB of GPU memory per model for training and fine-tuning. Simultaneously co-optimizing software and hardware using DS and expanding this supernetwork to include all possible hardware combinations would require an excessive amount of GPU resources for training and convergence \cite{elsken2019neural, poyser2024neural, guo2020breaking, krestinskaya2024neural}.

EA, in contrast, scales linearly with generations and supports parallel evaluation of populations. It is derivative-free, handles discrete parameters without relaxation, and achieves good coverage via crossover and mutation. Typical configurations use 30–100 generations with populations of 50–150. EA’s diversity helps escape local minima and converge efficiently, even in large search spaces \cite{elsken2019neural, poyser2024neural, guo2020breaking, krestinskaya2024neural}.
Therefore, EA is selected for CIMNAS due to its scalability, parallelism, and compatibility with discrete hardware-aware search. Additionally, we use a pre-trained supernetwork-based accuracy predictor, similar to DS, to efficiently estimate candidate performance.

}

\subsubsection{{\color{black} CIMNAS algorithm}}

Preliminary experiments showed that unconstrained optimization often results in CIM designs with infeasibly large on-chip areas. To address this, CIMNAS focuses on area-constrained optimization, which serves two main purposes: it excludes designs with excessively large areas that are impractical to fabricate as a single chip, and it drives the evolutionary algorithm (EA) to converge more quickly toward reasonably sized CIM hardware compared to unconstrained optimization. The network score of a given CIM-based neural network design, $\alpha$, is calculated using an objective function, $f$, that incorporates energy $E_{\alpha}$, delay $D_{\alpha}$ (latency across all layers), on-chip area $A_{\alpha}$, and predicted performance accuracy $Acc_{\alpha}$ for the sampled design:

\begin{equation}\label{eq:o2}
\begin{aligned}
f_{\alpha} = f (E_{\alpha}, D_{\alpha}, A_{\alpha}, Acc_{\alpha}) \\
\mathrm{s.t.} \, A_{\alpha} \leq A_{constr} \;
\end{aligned}
\end{equation}
where $A_{constr}$ represents the area constraint. We focus on EDAP optimization by minimizing the objective function $f_{\alpha} = \frac{E_{\alpha} \times D_{\alpha} \times A_{\alpha}}{Acc_{\alpha}}$. Accuracy is included directly in the objective function to ensure that EDAP optimization does not compromise performance accuracy. We avoid setting accuracy as a constraint because the search is based on a pre-trained supernetwork, which inherently prevents the sampling of low-accuracy designs, eliminating the need to filter out designs with insufficient accuracy. The most compact supernetwork configuration—comprising 2 bottleneck blocks per bottleneck layer, a kernel size of 3, and an expansion factor of 4 in all layers—achieves 64.9\% accuracy on the ImageNet dataset when using the lowest bit precision of 4 bits for both weights and inputs. Additionally, we incorporate on-chip area as both an objective and a constraint to ensure that the algorithm not only minimizes area but also avoids converging to designs with unreasonably large on-chip areas, which often tend to reduce energy consumption.
In Section \ref{results}, we present experiments with other area-constrained objective functions for comparison.

During the initial sampling to construct the starting population for the EA, design samples that cannot accommodate the neural network model are removed (Algorithm \ref{alg}). As a result, the initial population consists only of feasible designs, while any samples that do not meet the initial design constraints are discarded. This approach reduces the presence of infeasible designs in the search space and minimizes the likelihood of EA mutations generating and propagating unfit designs throughout the search process. In addition, if an infeasible design sample is produced after crossover and mutation, it is discarded from the population in the new generation.

After initializing the population (Algorithm \ref{alg}), the evolutionary algorithm is executed over $G$ generations. For each sampled design, the full precision weights and activations from the supernetwork are quantized to generate the corresponding histograms. Hardware metrics are then evaluated, and design accuracy is estimated using an accuracy predictor. The design samples are subsequently ranked based on the objective function defined in Eq. \ref{eq:o2}. Selected designs undergo binary crossover and polynomial mutation \cite{deb2007self, blank2020pymoo}, with crossover and mutation probabilities set to $\mathbb{P}_c = 0.95$ and $\mathbb{P}_m = 0.95$, respectively. The distribution indices for these operations are $\eta_c = 3$ and $\eta_m = 3$, values within the typical range of 3 to 30. These parameters promote exploration and maintain population diversity, enabling a broad search of the design space. Following mutation, the evaluation process is repeated for the newly generated population. After each generation, all design samples and their associated metrics are stored. The final optimized designs are selected from the entire set of stored samples based on the objective function in Eq. \ref{eq:o2}. This approach not only ensures comprehensive optimization but also avoids redundant evaluations of identical samples, thereby reducing computational overhead and search time.

\subsection{Evaluation of software and hardware metrics}
\label{sec:C}

\subsubsection{{ \color{black}Performance accuracy evaluation and accuracy predictor}}
\label{accuracy_eval}

For software metrics evaluation, we use the pre-trained full-precision "Once-for-all" supernetwork from \cite{cai2019once} and an accuracy predictor for quantized designs from \cite{wang2020apq}. This accuracy predictor is trained by sampling models from the supernetwork, quantizing them, and performing a single epoch of quantization-aware fine-tuning.
{\color{black}
The accuracy predictor is a three-layer neural network with hidden layers of size 400.
First, a full-precision accuracy predictor is trained. The input to this predictor consists of all neural network model parameters from the search space illustrated in Fig. \ref{f2}, encoded using a one-hot encoding scheme. The output is the predicted performance accuracy corresponding to the given model parameters. The full-precision predictor is trained using 80,000 samples of neural network configurations and their corresponding performance accuracies.
Once the full-precision predictor is trained, it is extended to support quantization parameters from the quantization search space, which are also encoded using one-hot encoding. This quantized accuracy predictor is derived by expanding the input layer of the full-precision model to include the quantization policy representations and fine-tuning this predictor using 2,500 sampled neural network architectures, each evaluated with 10 different quantization policy configurations. The quantized accuracy predictor outputs the predicted performance accuracy of a quantized neural network, considering both the sampled model, and quantization parameters \cite{wang2020apq}.
}

The accuracy predictor greatly reduces search time; while it takes just 5 GPU seconds per network, directly fine-tuning each quantized design would require at least 4-5 GPU hours. Without the accuracy predictor, exploring such an extensive search space would be impractical. Although the accuracy predictor does not provide exact accuracy values, its relative predictions are consistent enough to compare sampled designs during the search. Final accuracy for the top-selected networks in Section \ref{results} is obtained through fine-tuning these models.

\subsubsection{{\color{black}Hardware metrics evaluation}}

{\color{black}
Hardware metrics evaluation is performed using the CiMLoop simulator \cite{andrulis2024cimloop}, a highly flexible, cycle-accurate simulation framework designed for architecture-level evaluation of CIM-based hardware accelerators. CiMLoop supports detailed modeling of architecture, circuit, and device parameters, making it well suited for evaluating diverse CIM-based design configurations. It integrates the Timeloop simulator \cite{parashar2019timeloop} for model-to-hardware mapping and employs the Accelergy framework \cite{wu2019accelergy} for energy estimation.
CiMLoop was selected for its relatively high simulation speed and accuracy close to that of NeuroSim \cite{peng2020dnn+}, while offering enhanced flexibility. Specifically, CiMLoop achieves an average error of approximately 3\% in hardware estimations \cite{andrulis2024cimloop}. In contrast to NeuroSim, CiMLoop supports parallel processing of model layers, significantly reducing runtime. For instance, while NeuroSim requires approximately 945 seconds to process a MobileNetV2 architecture on a single core, CiMLoop completes the same task in 220.5 seconds. When parallelized across 64 CPU cores, the runtime is further reduced to approximately 25 seconds. Therefore, we achieve almost 40$\mathbf{\times}$ faster hardware metrics estimation compared to NeuroSim.
This parallelism is especially advantageous in our CIMNAS framework, where evolutionary algorithm (EA) search generations and CiMLoop-based hardware evaluations are both parallelizable. This enables highly efficient large-scale search, fully utilizing multithreading capabilities. For a single EA generation, we achieve at least a 10$\mathbf{\times}$ speedup compared to using NeuroSim. The combination of fast simulation speed and low estimation error makes CiMLoop a practical and scalable choice for evaluating thousands of candidate architectures in CIMNAS.
}

Instead of directly processing the inputs and weights of each layer, as done in time-consuming cycle-accurate simulators that simulate every operation, CiMLoop leverages histograms of these values. This approach significantly reduces simulation time while maintaining reliable, data-dependent performance estimates.
As "Once-for-all" is a pre-trained full-precision supernetwork, we quantize the inputs, weights, and outputs of each layer and convert them into corresponding histograms for CiMLoop to ensure performance estimation based on realistic data. The histogram generation function was implemented as described in \cite{andrulis2024cimloop}. Histograms for the inputs and outputs of each layer are generated using a randomly selected image from the ImageNet dataset. Each design evaluation takes about 30 seconds on a 64-core CPU, allowing the search to proceed efficiently without additional speed-up methods, such as performance predictors used in \cite{sun2023gibbon}.

\begin{table*}[t]
\caption{Comparison of CIMNAS with baseline and state-of-the-art search methods. The proposed framework is evaluated against a two-stage approach and an XPert-style search, following strategies from~\cite{han2024comn} and~\cite{moitra2023xpert}.}
\resizebox{2.0\columnwidth}{!}{%
\begin{tabular}{|lcccccccccccc|}
\hline
\multicolumn{1}{|l|}{\multirow{2}{*}{\textbf{\begin{tabular}[c]{@{}l@{}}Optimization\\ approach\end{tabular}}}} & \multicolumn{3}{c|}{\textbf{Accuracy}}                                                                             & \multicolumn{1}{c|}{\multirow{2}{*}{\textbf{\begin{tabular}[c]{@{}c@{}}Energy \\ (mJ)\end{tabular}}}} & \multicolumn{1}{c|}{\multirow{2}{*}{\textbf{\begin{tabular}[c]{@{}c@{}}Delay \\ (us)\end{tabular}}}} & \multicolumn{1}{c|}{\multirow{2}{*}{\textbf{\begin{tabular}[c]{@{}c@{}}Area\\ (mm$^2$)\end{tabular}}}} & \multicolumn{1}{c|}{\multirow{2}{*}{\textbf{\begin{tabular}[c]{@{}c@{}}EDAP \\ (mJ*ms*mm$^2$)\end{tabular}}}} & \multicolumn{1}{c|}{\multirow{2}{*}{\textbf{\begin{tabular}[c]{@{}c@{}}Search \\ score\end{tabular}}}}       & \multicolumn{1}{c|}{\multirow{2}{*}
{\textbf{\begin{tabular}[c]{@{}c@{}}TOPS/\\W\end{tabular}}}}
& \multicolumn{1}{c|}{\multirow{2}{*}{\textbf{\begin{tabular}[c]{@{}c@{}}TOPS/\\mm$^2$\end{tabular}}}}

& \multicolumn{1}{c|}{\multirow{2}{*}{\textbf{\begin{tabular}[c]{@{}c@{}}Hardware \\ utilization\end{tabular}}}} & \multirow{2}{*}{\textbf{\begin{tabular}[c]{@{}c@{}}Design\\ diversity$^{*2}$\end{tabular}}} \\ \cline{2-4}
\multicolumn{1}{|l|}{}                                                                                          & \multicolumn{1}{c|}{\textbf{Top-1}} & \multicolumn{1}{c|}{\textbf{Top-5}} & \multicolumn{1}{c|}{\textbf{Change$^{*1}$}} & \multicolumn{1}{c|}{}                                                                                 & \multicolumn{1}{c|}{}                                                                                & \multicolumn{1}{c|}{}                                                                               & \multicolumn{1}{c|}{}                                                                                      & \multicolumn{1}{c|}{}                                                                                        & \multicolumn{1}{c|}{}                                 & \multicolumn{1}{c|}{}                                   & \multicolumn{1}{c|}{}                                                                                          &                                                                                        \\ \hline
\multicolumn{1}{|l|}{\textbf{\begin{tabular}[c]{@{}l@{}}Baseline 1 \\  (median of the \\parameters)\end{tabular}}}                                                                       & \multicolumn{1}{c|}{73.00\%}        & \multicolumn{1}{c|}{91.20\%}        & \multicolumn{1}{c|}{-}                 & \multicolumn{1}{c|}{0.95}                                                                           & \multicolumn{1}{c|}{6.94}                                                                          & \multicolumn{1}{c|}{3691}                                                                        & \multicolumn{1}{c|}{24.33}                                                                                 & \multicolumn{1}{c|}{\begin{tabular}[c]{@{}c@{}}EDAP/Acc: 0.34\\ L/Acc: 95.1\\ EA/Acc: 48.02\end{tabular}}  & \multicolumn{1}{c|}{1.36}                             & \multicolumn{1}{c|}{0.71}                               & \multicolumn{1}{c|}{0.26}                                                                                      & -                                                                                      \\ \hline
\multicolumn{1}{|l|}{\textbf{\begin{tabular}[c]{@{}l@{}}Baseline 2 \\  (random 1000)\end{tabular}}}                 & \multicolumn{1}{c|}{73.00\%}        & \multicolumn{1}{c|}{91.20\%}        & \multicolumn{1}{c|}{-}                 & \multicolumn{1}{c|}{1.15}                                                                           & \multicolumn{1}{c|}{3.80}                                                                          & \multicolumn{1}{c|}{14033}                                                                       & \multicolumn{1}{c|}{28.19$^{*3}$}                                                                               & \multicolumn{1}{c|}{\begin{tabular}[c]{@{}c@{}}EDAP/Acc: 0.38\\ L/Acc: 52.1\\ EA/Acc: 237.19\end{tabular}} & \multicolumn{1}{c|}{1.4}                              & \multicolumn{1}{c|}{0.8}                                & \multicolumn{1}{c|}{0.35}                                                                                      & -                                                                                      \\ \hline
\multicolumn{13}{|c|}{\textbf{EDAP - focused optimization: search score EDAP/Acc (mJ*ms*mm$^2$/\%)}}                                                                                                                                                                                                                                                                                                                                                                                                                                                                                                                                                                                                                                                                                                                                                                                                                                                                                                                                                                                                                            \\ \hline
\multicolumn{1}{|l|}{\textbf{Two-stage search}}                                                                 & \multicolumn{1}{c|}{74.25\%}        & \multicolumn{1}{c|}{92.01\%}        & \multicolumn{1}{c|}{$\uparrow$ 1.25\%}            & \multicolumn{1}{c|}{0.43}                                                                           & \multicolumn{1}{c|}{8.80}                                                                          & \multicolumn{1}{c|}{465}                                                                         & \multicolumn{1}{c|}{1.76}                                                                                  & \multicolumn{1}{c|}{0.0250}                                                                                & \multicolumn{1}{c|}{8.41}                             & \multicolumn{1}{c|}{2.42}                               & \multicolumn{1}{c|}{0.53}                                                                                      & medium                                                                                 \\ \hline
\multicolumn{1}{|l|}{\textbf{XPert-like search}}                                                                & \multicolumn{1}{c|}{69.60\%}        & \multicolumn{1}{c|}{88.90\%}        & \multicolumn{1}{c|}{$\downarrow$ 3.4\%}             & \multicolumn{1}{c|}{0.27}                                                                           & \multicolumn{1}{c|}{2.72}                                                                          & \multicolumn{1}{c|}{235}                                                                         & \multicolumn{1}{c|}{0.17}                                                                                  & \multicolumn{1}{c|}{0.0025}                                                                                & \multicolumn{1}{c|}{6.56}                             & \multicolumn{1}{c|}{8.91}                               & \multicolumn{1}{c|}{0.68}                                                                                      & low                                                                                    \\ \hline
\multicolumn{1}{|l|}{\textbf{CIMNAS}}                                                                           & \multicolumn{1}{c|}{73.81\%}        & \multicolumn{1}{c|}{91.80\%}        & \multicolumn{1}{c|}{$\uparrow$ 0.81\%}            & \multicolumn{1}{c|}{0.33}                                                                           & \multicolumn{1}{c|}{3.55}                                                                          & \multicolumn{1}{c|}{234}                                                                         & \multicolumn{1}{c|}{0.27}                                                                                  & \multicolumn{1}{c|}{0.0037}                                                                                & \multicolumn{1}{c|}{6.56}                             & \multicolumn{1}{c|}{9.08}                               & \multicolumn{1}{c|}{0.60}                                                                                       & high                                                                                   \\ \hline
\multicolumn{13}{|c|}{\textbf{Latency - focused optimization: search score D/Acc (ms/\%)}}                                                                                                                                                                                                                                                                                                                                                                                                                                                                                                                                                                                                                                                                                                                                                                                                                                                                                                                                                                                                                               \\ \hline
\multicolumn{1}{|l|}{\textbf{Two-stage search}}                                                                 & \multicolumn{1}{c|}{74.25\%}        & \multicolumn{1}{c|}{92.01\%}        & \multicolumn{1}{c|}{$\uparrow$ 1.25\%}            & \multicolumn{1}{c|}{0.39}                                                                                 & \multicolumn{1}{c|}{2.10}                                                                                & \multicolumn{1}{c|}{797}                                                                               & \multicolumn{1}{c|}{0.65}                                                                                      & \multicolumn{1}{c|}{28.3}                                                                                        & \multicolumn{1}{c|}{9.41}                             & \multicolumn{1}{c|}{9.82}                               & \multicolumn{1}{c|}{0.53}                                                                                      & medium                                                                                 \\ \hline
\multicolumn{1}{|l|}{\textbf{XPert-like search}}                                                                & \multicolumn{1}{c|}{72.76\%}        & \multicolumn{1}{c|}{90.99\%}        & \multicolumn{1}{c|}{$\downarrow$ 0.24\%}            & \multicolumn{1}{c|}{0.27}                                                                           & \multicolumn{1}{c|}{1.40}                                                                          & \multicolumn{1}{c|}{464}                                                                         & \multicolumn{1}{c|}{0.18}                                                                                  & \multicolumn{1}{c|}{19.2}                                                                              & \multicolumn{1}{c|}{7.67}                             & \multicolumn{1}{c|}{13.55}                              & \multicolumn{1}{c|}{0.59}                                                                                      & low                                                                                    \\ \hline
\multicolumn{1}{|l|}{\textbf{CIMNAS}}                                                                           & \multicolumn{1}{c|}{73.71\%}        & \multicolumn{1}{c|}{91.74\%}        & \multicolumn{1}{c|}{$\uparrow$ 0.71\%}            & \multicolumn{1}{c|}{0.30}                                                                            & \multicolumn{1}{c|}{1.47}                                                                          & \multicolumn{1}{c|}{467}                                                                         & \multicolumn{1}{c|}{0.20}                                                                                   & \multicolumn{1}{c|}{19.9}                                                                              & \multicolumn{1}{c|}{6.45}                             & \multicolumn{1}{c|}{11.00}                                 & \multicolumn{1}{c|}{0.57}                                                                                      & high                                                                                   \\ \hline
\multicolumn{13}{|c|}{\textbf{Energy/area - focused optimization: search score EA/Acc (mJ*mm$^2$/\%) }}                                                                                                                                                                                                                                                                                                                                                                                                                                                                                                                                                                                                                                                                                                                                                                                                                                                                                                                                                                                                                               \\ \hline
\multicolumn{1}{|l|}{\textbf{Two-stage search}}                                                                 & \multicolumn{1}{c|}{74.25\%}        & \multicolumn{1}{c|}{92.01\%}        & \multicolumn{1}{c|}{$\uparrow$ 1.25\%}            & \multicolumn{1}{c|}{0.56}                                                                                 & \multicolumn{1}{c|}{2.14}                                                                                & \multicolumn{1}{c|}{472}                                                                               & \multicolumn{1}{c|}{0.57}                                                                                      & \multicolumn{1}{c|}{3.56}                                                                                        & \multicolumn{1}{c|}{5.92}                             & \multicolumn{1}{c|}{13.30}                               & \multicolumn{1}{c|}{0.31}                                                                                      & medium                                                                                 \\ \hline
\multicolumn{1}{|l|}{\textbf{XPert-like search}}                                                                & \multicolumn{1}{c|}{69.57\%}        & \multicolumn{1}{c|}{89.21\%}        & \multicolumn{1}{c|}{$\downarrow$ 3.43\%}            & \multicolumn{1}{c|}{0.20}                                                                           & \multicolumn{1}{c|}{1.21}                                                                          & \multicolumn{1}{c|}{799}                                                                         & \multicolumn{1}{c|}{0.19}                                                                                  & \multicolumn{1}{c|}{3.44}                                                                                    & \multicolumn{1}{c|}{8.79}                             & \multicolumn{1}{c|}{8.48}                               & \multicolumn{1}{c|}{0.30}                                                                                       & low                                                                                    \\ \hline
\multicolumn{1}{|l|}{\textbf{CIMNAS}}                                                                           & \multicolumn{1}{c|}{71.85\%}        & \multicolumn{1}{c|}{90.58\%}        & \multicolumn{1}{c|}{$\downarrow$ 1.5\%}             & \multicolumn{1}{c|}{0.22}                                                                           & \multicolumn{1}{c|}{2.90}                                                                          & \multicolumn{1}{c|}{234}                                                                         & \multicolumn{1}{c|}{0.14}                                                                                  & \multicolumn{1}{c|}{0.71}                                                                                    & \multicolumn{1}{c|}{6.75}                             & \multicolumn{1}{c|}{1.63}                               & \multicolumn{1}{c|}{0.52}                                                                                      & high                                                                                   \\ \hline
\multicolumn{13}{|l|}{\begin{tabular}[c]{@{}l@{}}$^{*1}$: accuracy change compared to the baseline top-1, $\downarrow$ - accuracy decreased, $\uparrow$ - accuracy increased. $^{*2}$: In the context of a large search space with several  potential optima, \\design diversity reflects the degree of uniqueness among the top five optimized designs. $^{*3}$: average across 1000 randomly sampled hardware configurations \end{tabular}}                                                                                                                                                                                                                                                                                                                                                                                                                                                                                                                                                                                                                                                                                                                                                                                                                                                                                                                     \\ \hline
\end{tabular}
}
\label{t2}
\end{table*}

\begin{figure*}[t!]
    \includegraphics[width=\textwidth]{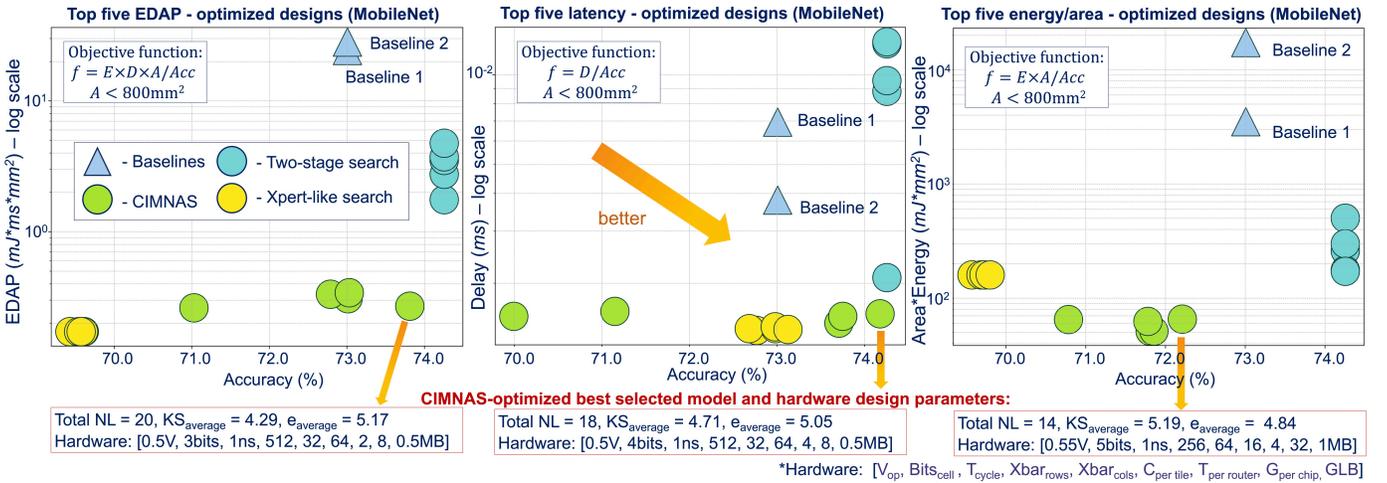}
    \caption{Trade-offs between hardware performance and accuracy for the top 5 selected designs, along with the parameters of the highest-scoring CIMNAS-selected designs for each experiment.}
    \label{f4}
\end{figure*}

\subsection{{\color{black} Computational complexity, memory cost, and runtime of CIMNAS}}

{\color{black} 
The computational complexity of CIMNAS is determined by two main components: the EA-based search and the supernetwork-based accuracy predictor training. The complexity of the EA search is outlined in Table II, and is primarily governed by the number of generations and the population size. Importantly, EA scales linearly with the number of hyperparameters per generation and does not suffer from exponential growth. Within each generation, candidate evaluations can be fully parallelized. Moreover, the EA search has low computational overhead and does not require GPU resources, making it suitable for parallel execution across multiple CPU cores. GPU resources are mainly utilized to accelerate the accuracy evaluations performed by the supernetwork-based predictor.
The training of the accuracy predictor, based on a supernetwork, is more computationally demanding. It requires substantial GPU memory and compute resources for both initial training and fine-tuning to support quantization (as described in Section \ref{accuracy_eval}). However, once trained, this predictor can be reused across multiple hardware configurations, enabling efficient evaluation of diverse hardware–software co-design scenarios without retraining. 
}

\section{Results}
\label{results}

\subsection{{\color{black} Simulation setup}}
\label{resultsetup}

CIMNAS searches were conducted with a population size of $P=150$ architectures over $G=70$ generations. Each search takes approximately 1.75 to 2.5 days, utilizing 64 CPU cores for parallel layer-wise hardware performance evaluation and a single GPU for the accuracy predictor. The initial sampling phase requires 8–10 minutes, while each generation takes around 50 minutes to complete. Hardware evaluations were performed using 32nm CMOS technology and RRAM devices from \cite{lu2021neurosim}, with an area constraint of $A_{constr} = 800 \text{ mm}^2$, reflecting a reasonable die size for single-chip fabrication \cite{choquette20213}.
To ensure fair comparison across search techniques, we fixed the initial seed to start each search with a similar initial population. Accuracy was evaluated on the ImageNet dataset with 1000 classes. Since accuracy evaluation during the search relies on the accuracy predictor, the final accuracy for each quantized architecture is obtained by fine-tuning the quantized model.

In this work, we define two primary baselines for comparison, differing in hardware parameters. In both baseline architectures, the software model chosen for reference is a standard-size 8-bit MobileNetV2 \cite{sandler2018mobilenetv2} neural network. It should be noted that the search space includes configurations with larger kernel sizes and additional layers, allowing NAS to potentially discover networks with higher performance accuracy than the baseline.
Baseline 1 is constructed by sampling the median value of each hardware parameter, with specifications: $\mathrm{V_{op}}=0.7V$, $\mathrm{Bits_{cell}}=4$, $\mathrm{T_{cycle}}=4ns$, $\mathrm{Xbar_{rows}}=\mathrm{Xbar_{cols}}=256$, $\mathrm{C_{per \, tile}}=16$, $\mathrm{T_{per \, router}}=8$, $\mathrm{G_{per \, chip}}=16$, and $\mathrm{GLB}=4$ MB.
To ensure unbiased representation, a second baseline is defined by randomly sampling 1000 hardware configurations from the search space and calculating the mean performance metrics, providing a fair representation of the search space.

We test three optimization cases with CIMNAS: (1) EDAP and accuracy, (2) delay and accuracy, and (3) energy, area, and accuracy. The corresponding objective functions for the CIMNAS search are $f = \frac{E \times D \times A}{Acc}$, $f = \frac{D}{Acc}$, and $f = \frac{E \times A}{Acc}$.
We compare CIMNAS with two methods: two-stage search and XPert-like search. The two-stage search approach sequentially optimizes software and precision parameters for high accuracy without hardware considerations, followed by CIM hardware optimization in the second stage \cite{krestinskaya2024neural}, this approach is similar to the one in \cite{han2024comn}.
The XPert-like search method, based on \cite{moitra2023xpert}, also uses a two-stage optimization, with the first stage focusing on latency and area, and the second stage targeting energy and accuracy. For EDAP and accuracy optimization, the first stage focuses on model and hardware parameters to optimize area and delay, followed by quantization parameter optimization in the second stage for accuracy and energy efficiency. For delay and accuracy, latency is optimized in the first stage, with accuracy refinement in the second. For energy, area, and accuracy, area is optimized initially, followed by energy and accuracy in the second stage.

\subsection{{\color{black} CIMNAS simulation results}}

Table \ref{t2} and Fig. \ref{f4} show the simulation results and a comparison of the proposed methods with baseline and state-of-the-art approaches. 
For EDAP optimization, CIMNAS achieves a reduction of $90.1\times$ to $104.5\times$ in the energy-delay-area product, along with a $4.68\times$ to $4.82\times$ increase in energy efficiency, an $11.35\times$ to $12.78\times$ improvement in area efficiency, a $1.7\times$ to $2.3\times$ boost in hardware utilization, and a $0.81\%$ gain in performance accuracy compared to the baseline architectures.
With EDAP optimization, CIMNAS identifies architectures that reduce hardware metrics without compromising performance accuracy.
Latency-focused optimization follows a similar trend, reducing delay by a factor of $2.5\times$ to $4.7\times$ compared to the baseline while increasing performance accuracy by $0.71\%$. For energy-area optimization, CIMNAS achieves a $68\times$ to $313\times$ improvement in the $E \times A$ score, with only a $1.5\%$ accuracy reduction—significantly lower than the $3.43\%$ drop seen with XPert-like search.

In contrast, the two-stage search achieves the highest accuracy by optimizing software parameters first to maximize performance; however, it fails to reduce EDAP and optimize hardware efficiency. The XPert-like search achieves superior hardware performance by minimizing latency and/or area in the first stage, but it cannot attain high accuracy afterward due to hardware restrictions imposed in the initial stage. Consequently, CIMNAS achieves the best balance between accuracy and hardware metrics. This trend is evident in Fig. \ref{f4}, which shows the trade-offs between hardware metrics and performance accuracy for the top five selected designs from each of the three approaches.
The same trends are observed in the other two experiments focused on latency and on energy/area optimization.

Given the large search space used in these experiments, multiple optimal parameter combinations with similar scores are possible. To assess this, we evaluate the design diversity of each method, where design diversity indicates how well a search algorithm captures various high-scoring combinations. 
Fig. \ref{f4} shows how close the top five designs are to each other for each approach. 
The XPert-like search yields the least diverse architectures, as it optimizes model and hardware parameters in the first stage, and the subsequent quantization policy optimization in the second stage does not generate a wide variety of designs. The two-stage approach exhibits slightly more diversity since hardware parameters are adjusted in the second stage.
Both the two-stage and XPert-like approaches tend to fall into local minima and fail to achieve the best balance between accuracy and hardware performance. 
Capturing a diverse set of optimal designs within the search space is important for HW-NAS, as it benefits later design stages, such as transistor-level simulations, layout development, and the fabrication of the final CIM chip.

\begin{figure}[t!]
    \centering
    \includegraphics[width=\columnwidth]{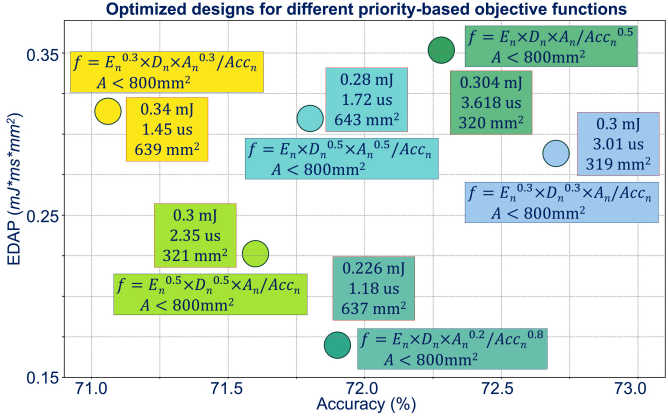}
    \caption{Demonstration of priority-based optimization.}
    \label{f5}
\end{figure}

\subsection{{\color{black} Priority-based optimization}}

For hardware design, it is crucial to evaluate trade-offs carefully, as one hardware metric can be more critical than another when optimizing system performance.
Therefore, we also tested the proposed framework for the priority-based objective function shown in Fig. \ref{f5}. In priority-based optimization, the objective function $f = \frac{E^{a}_{n} \times D^{b}_{n} \times A^{c}_{n}}{Acc^{d}_{n}}$ uses the hardware metrics $E_{n}$, $D_{n}$, $A_{n}$, and accuracy $Acc_{n}$ normalized to the values of the first obtained sample, where $a$, $b$, $c$, and $d$ are priority coefficients in the range of 0 to 1. Depending on the optimization priorities, hardware designs with similar EDAP can exhibit distinct performance in terms of hardware metrics; for example, on-chip area and accuracy-focused optimization with objective function $f = \frac{E^{0.3}_{n} \times D^{0.3}_{n} \times A_{n}}{Acc_{n}}$ favors designs with reduced area and enhanced accuracy, while energy and delay-focused optimization with objective function $f = \frac{E_{n} \times D_{n} \times A^{0.2}_{n}}{Acc^{0.8}_{n}}$ targets designs with minimized energy consumption and delay. In addition, we demonstrate that the search is also sensitive to the objective coefficients. For example, in an area and accuracy-focused optimization setting, $a=b=0.3$ for energy and delay priorities leads to designs with enhanced accuracy and smaller on-chip area compared to when $a=b=0.5$. This highlights the importance of fine-tuning priority coefficients to achieve optimal hardware design outcomes based on specific performance goals.

\begin{figure}[t!]
    \centering
    \includegraphics[width=1.0\columnwidth]{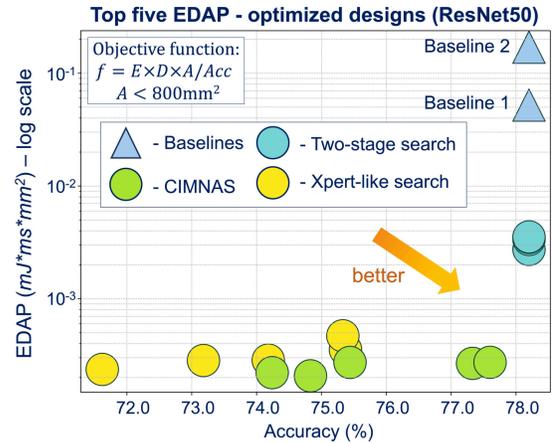}
    \caption{{\color{black}(a) Search space and (b) hardware system configuration of SRAM-based ResNet50 architecture. (c) Performance demonstration of the proposed CIMNAS framework applied to the ResNet50 model within the SRAM-based CIM design space.}}
    \label{fresnet}
\end{figure}

\subsection{{\color{black} Adaptability and robustness of the framework}}
\label{resultsAdapt}

{{\color{black}

To demonstrate the adaptability of the CIMNAS framework and the robustness of its results beyond MobileNet and the RRAM-based weight-stationary CIM design, we extend our evaluation to a different neural network architecture, a new CIM hardware configuration, and a different CMOS technology node. Specifically, we apply CIMNAS to the ResNet50 architecture using an SRAM-based CIM design with weight swapping, evaluated at the 7nm CMOS technology node.

In this experiment, a new model search space, illustrated in Fig. \ref{fresnet}(a), is based on the ResNet50 architecture. The model search space includes the depth of the bottleneck blocks, which defines how many times each block is repeated. It also includes a width multiplier that scales the number of channels across the network and an expansion factor that controls the ratio between the number of channels in the intermediate (expanded) layer versus the input/output channels in each bottleneck block.
The quantization search space includes precision settings for both normal and pointwise convolutions within the bottleneck layers, while the input convolution and output linear layers are fixed to 8-bit precision. The full-precision accuracy predictor is derived from a supernetwork trained on all parameter configurations within this ResNet50-based search space \cite{cai2019once}. The quantized accuracy predictor is then obtained by fine-tuning the full-precision accuracy predictor for the quantization search space shown in Fig.\ref{fresnet}(a).

The SRAM-based hardware configuration is shown in Fig. \ref{fresnet}(b). Unlike the RRAM-based weight-stationary architecture used in Fig. \ref{f2}, which requires all neural network weights to fit on-chip, the SRAM-based design in this experiment allows weight swapping. This enables sequential processing of layer groups on the same hardware: initial layers are executed first, followed by loading the next set of weights from DRAM, and so on until the final layer. The simulation includes the energy and latency overheads associated with transferring weights from DRAM to the chip. A standard 6T SRAM cell is used as in \cite{khwa201865nm}. The hardware search space is similar to the one in previous RRAM-based experiments, but slightly extended to support the larger architecture (Fig.\ref{fresnet}(a)).

The performance of CIMNAS on this ResNet50 search space using a 7nm SRAM-based CIM system is shown in Fig.\ref{fresnet}(c). Two baseline architectures are defined following the same methodology as in Section \ref{resultsetup}, both being conventional 8-bit ResNet50 models. For the hardware parameters, Baseline 1 represents a median configuration with $\mathrm{V_{op}}=0.7$V, $\mathrm{T_{cycle}}=4$ns, $\mathrm{Xbar_{rows}}=\mathrm{Xbar_{cols}}=128$, $\mathrm{C_{per\, tile}}=32$, $\mathrm{T_{per\, router}}=8$, $\mathrm{G_{per\, chip}}=32$, and $\mathrm{GLB}=4$MB. Baseline 2 reflects the mean hardware performance across 1,000 randomly sampled configurations from the search space.

Fig.\ref{fresnet}(c) compares CIMNAS with the other state-of-the-art frameworks. The 8-bit ResNet50 baseline achieves a Top-1 accuracy of 78.2\% and a Top-5 accuracy of 91.8\%. The best architecture discovered by CIMNAS reaches 77.6\% Top-1 accuracy and 91.2\% Top-5 accuracy, with only a 0.6\% drop, an insignificant degradation, while achieving significant EDAP improvements. Specifically, CIMNAS achieves 251.1$\times$ and 819.5$\times$ reductions in EDAP compared to Baseline 1 and Baseline 2, respectively.
The results in Fig.\ref{fresnet}(c) show that CIMNAS outperforms two-stage methods and Xpert-like approaches in this new setting. While the accuracy remains close to the baseline, the substantial gains in EDAP are due to the fact that the search space is dominated by hardware parameters, which have a greater impact on the hardware performance and EDAP optimization outcome, compared to the model or quantization parameters.

This experiment confirms that CIMNAS is highly adaptable to different neural network models, hardware configurations, and technology nodes, while maintaining strong performance. It demonstrates CIMNAS's ability to generalize and consistently outperform both baselines and state-of-the-art alternatives.

}
\section{Discussion}
\label{discussion}

\begin{figure}[t!]
    \centering
    \includegraphics[width=\columnwidth]{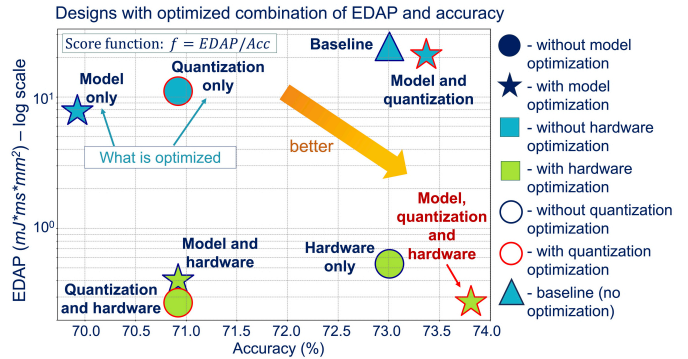}
    \caption{Impact of parameter co-optimization on hardware efficiency and performance accuracy. Designs are generated using CIMNAS with an EDAP/Accuracy objective function (minimizing EDAP while maximizing accuracy) {\color{black}for MobileNet search space.} The details of the baseline comparison are discussed in Section \ref{results}.}
    \label{f1}
\end{figure}

The results presented in this work highlight the critical importance of performing joint optimization across the model, quantization, and CIM hardware design spaces. 
In addition, 
Fig.\ref{f1} highlights the importance of an integrated model-quantization-hardware co-optimization to achieve optimal hardware efficiency and performance accuracy.
The graph represents seven different CIMNAS-based optimization experiments, each aimed at identifying CIM-based neural network implementations with an optimal combination of energy-delay-area product (EDAP) and performance accuracy. These experiments involved various parameter configurations, demonstrating that optimizing hardware parameters plays a significant role in enhancing hardware performance. The results indicate that focusing solely on neural network model optimization, even with hardware feedback, is insufficient to achieve a hardware-efficient solution. To obtain the optimal combination of EDAP and performance accuracy, it is essential to co-optimize model, quantization, and hardware parameters.

Moreover, traditional sequential or isolated optimization approaches often lead to suboptimal trade-offs, as decisions made in one domain (e.g., model architecture) can significantly impact the effectiveness and efficiency of others (e.g., hardware implementation). By contrast, CIMNAS enables a holistic co-design process, capturing complex interdependencies between the algorithmic and hardware parameters to find globally optimal configurations.
This joint optimization is especially critical for CIM systems, where quantization settings, memory cell behavior, circuit-level constraints, architecture-level design choices, and model performance are tightly interdependent and governed by complex non-linear relationships.
CIMNAS’s ability to explore an extremely large and diverse search space (on the order of $9.9\times10^{85}$ configurations) while still identifying EDAP-efficient and accurate solutions demonstrates its scalability and robustness.

CIMNAS serves as an initial tool in CIM hardware design automation, supporting the creation of energy- and area-efficient CIM chips for AI applications. As illustrated in Fig.\ref{f1} and Fig.\ref{f5}, the framework also supports flexible optimization modes, including the ability to optimize specific groups of parameters independently or to perform priority-based optimization, where certain design objectives are prioritized over others based on application needs.
{\color{black}
Section \ref{resultsAdapt} demonstrates the adaptability of CIMNAS to various neural network architectures, CIM hardware configurations, and CMOS technology nodes. The framework can support diverse design settings, including both weight-stationary and weight-swapping memory schemes, without requiring fundamental modifications. Moreover, CIMNAS shows strong robustness, maintaining effective optimization performance across search spaces dominated by different factors, whether hardware, model, or quantization parameters, highlighting its generalizability across different design scenarios.}
Looking ahead, CIMNAS can serve as a general-purpose framework for the automated co-design of energy-efficient edge AI systems. It can be extended to support other emerging memory technologies (e.g., FeFETs, RRAM), new quantization schemes (e.g., mixed-precision, non-uniform quantization), larger neural network models, or task-specific model constraints (e.g., latency-bound or memory-limited designs).
By bridging the gap between algorithm design and hardware realization, CIMNAS lays the foundation for next-generation CIM-aware neural architecture search frameworks that are adaptable, scalable, and efficient.

\section{Conclusion}
\label{conclusion}

We introduced CIMNAS, a CIM-aware NAS framework that jointly optimizes model, quantization, and hardware parameters across a comprehensive search space, including device-, circuit-, and architecture-level CIM hardware configurations.
We introduced CIMNAS, a CIM-aware NAS framework that jointly optimizes model, quantization, and hardware parameters across a comprehensive search space, including device-, circuit-, and architecture-level CIM hardware configurations.
CIMNAS achieves an optimal balance of hardware efficiency and performance without sacrificing accuracy, producing a diverse set of model-quantization-hardware parameter combinations.
{\color{black}For RRAM-based MobileNet architecture,} CIMNAS achieves up to a 104.5$\times$ reduction in EDAP, and improvements of 4.82$\times$ and 12.78$\times$ in energy and area efficiency, respectively, compared to the selected baseline. {\color{black} While for SRAM-based ResNet50 architecture, CIMNAS achieves an even greater EDAP reduction of up to 819.5$\times$. }
As part of future work, we aim to extend the algorithm to support a broader range of workloads and tasks beyond image classification. Additionally, we plan to develop prediction models to adapt search results to different hardware technologies, eliminating the need to rerun the search when migrating to new hardware and a new technology node.


\bibliographystyle{ieeetr}
\bibliography{references.bib}

\end{document}